\documentclass{article}
\usepackage[preprint]{spconf}
\usepackage{amsmath,epsfig}
\usepackage{url}
\usepackage[hidelinks]{hyperref}

\let\OLDthebibliography\thebibliography
\renewcommand\thebibliography[1]{
  \OLDthebibliography{#1}
  \setlength{\parskip}{0pt}
  \setlength{\itemsep}{0pt plus 0.3ex}
}

\title{FVV Live: real-time, low-cost, free viewpoint video}
\name{\begin{tabular}{c}Daniel~Berj\'on, Pablo~Carballeira\sthanks{P. Carballeira is now with Video Processing and Understanding Lab, Escuela Polit\'ecnica Superior, Universidad Aut\'onoma de Madrid, ES-28049, Madrid, Spain}, Juli\'an~Cabrera, Carlos~Carmona, \\
Daniel~Corregidor, C\'esar~D\'iaz, Francisco~Mor\'an, and Narciso~Garc\'ia\end{tabular}
\thanks{This work has been partially supported by the Ministerio de Ciencia, Innovaci\'on 
y Universidades (AEI/FEDER) of the Spanish Government under
project TEC2016-75981 (IVME) and by Huawei Technologies Co., Ltd.}
\thanks{\copyright~2020 IEEE. Personal use of this material is permitted.  Permission from IEEE must be obtained for all other uses, in any current or future media, including reprinting/republishing this material for advertising or promotional purposes, creating new collective works, for resale or redistribution to servers or lists, or reuse of any copyrighted component of this work in other works. \href{https://doi.org/10.1109/ICMEW46912.2020.9105977}{DOI: 10.1109/ICMEW46912.2020.9105977}}}
\address{Grupo de Tratamiento de Im\'agenes,
Information Processing and Telecommunications Center\\
Universidad Polit\'ecnica de Madrid, ES-28040, Madrid, Spain}

\copyrightnotice{978-1-7281-1485-9/20/\$31.00~\copyright2020 IEEE}

\begin{document}\sloppy
\ninept

\def\x{{\mathbf x}}
\def\L{{\cal L}}

\maketitle

\begin{abstract}
FVV Live is a novel real-time, low-latency, end-to-end free viewpoint
system including capture, transmission, synthesis on an edge server
and visualization and control on a mobile terminal. The system has
been specially designed for low-cost and real-time operation, only
using off-the-shelf components.
\end{abstract}
\section{Introduction}
\label{sec:intro}
Immersive video technologies have experienced significant development
over the past few years. One such technology is Free Viewpoint Video
(FVV), which enables users to navigate a scene by placing a virtual
camera ideally at any viewpoint of their choosing. This functionality
has the potential to improve user experience in many applications
such as broadcasting of any kind of events (sports, performances,
etc.) or interactive video communications (telepresence).

As it is typical of 3D reconstruction systems, FVV systems need to
see the 3D scene from many reference viewpoints (i.e., cameras) and
integrate the information from them to synthesize a model or virtual
view of the scene. However, unlike static 3D reconstruction applications,
FVV systems need to simultaneously capture many synchronized video
streams, which means that the sheer amount of input data easily exceeds
the capabilities of any single computer. Therefore, FVV systems are
inherently distributed, they need distinct nodes devoted to capture
and synthesis, and some means to efficiently encode and transmit data
among them.

Thus, the design and development of FVV systems presents a number
of challenges regarding video quality, cost and other functional requirements
such as real-time operation, which are frequently at odds with one
another. Clearly, high video quality benefits from having more and
higher resolution cameras, but that increases the cost and the amount
of data to be transmitted and processed, making it (even) more difficult
to meet real-time constraints, which is already hard with few cameras
due to the high computational complexity of typical view synthesis
algorithms. In fact, expensive systems in commercial operation such
as Intel True View~\cite{trueview} or 4DReplay~\cite{4dreplay}
have been designed to maximize visual quality but at the cost of sacrificing
real-time operation despite using tens of processing nodes.

In this paper, we present FVV Live, a novel end-to-end real-time, low-cost, FVV system that coordinates several nodes covering the various stages of the system, from capture to transmission, synthesis, and finally visualisation and control from a mobile terminal. FVV Live was conceived from the outset with the two key constraints of low-cost, only resorting to consumer-grade electronics, and real-time operation. These two constraints have informed many of the decisions taken during the design and development of the system. While other systems have been proposed in the literature or in the industry, FVV Live remains, to our knowledge, the only system operating in real-time and, furthermore, at a substantially lower cost than other proposals.

\begin{figure}[tbh]
\begin{centering}
\includegraphics[viewport=0bp 150bp 2048bp 1100bp,clip,width=0.85\columnwidth]{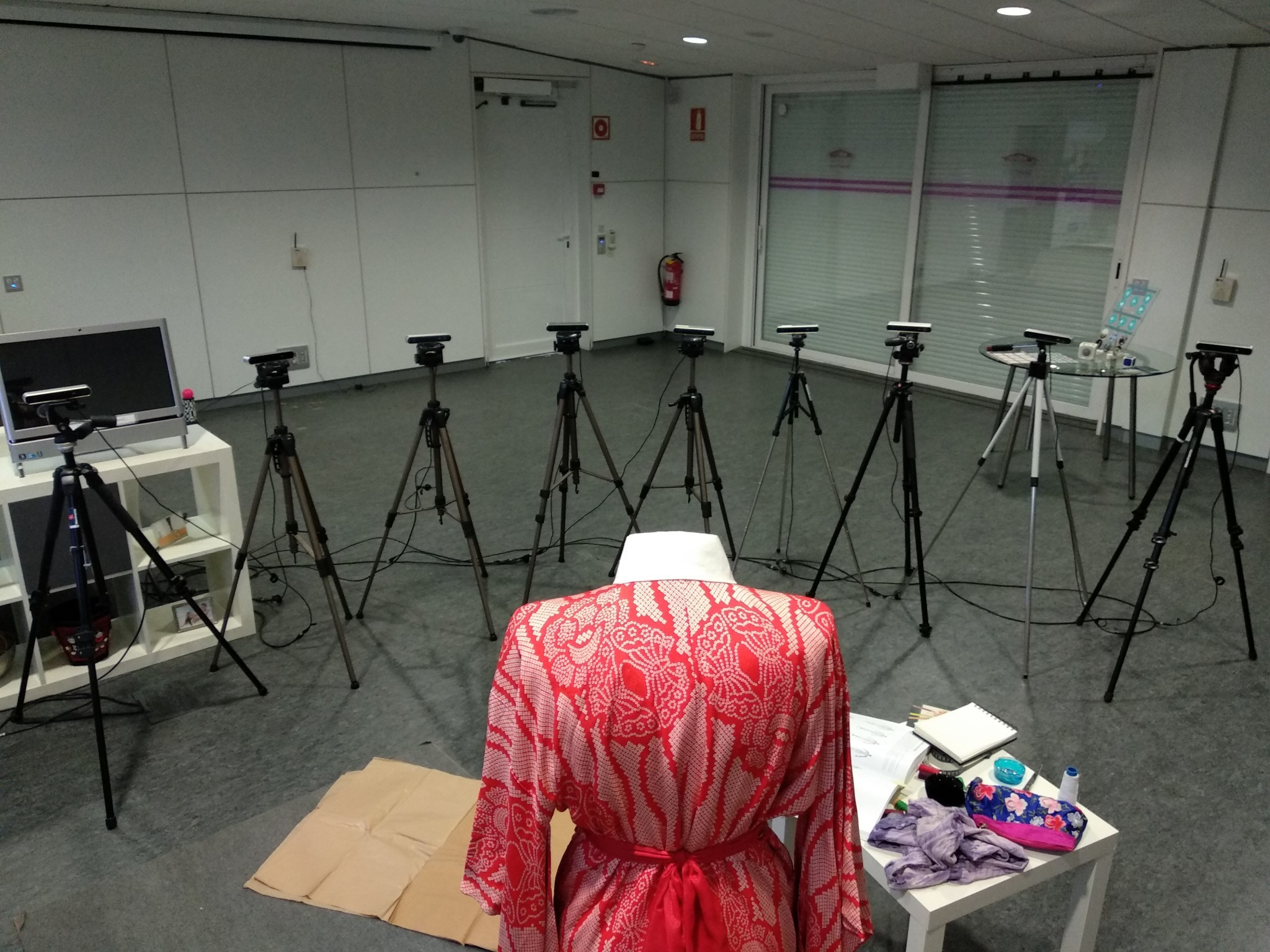}
\par\end{centering}
\caption{\label{fig:capture-setup}Overview of the capture setup.}
\end{figure}

\section{\label{sec:Capture-and-transmission}Capture and transmission}

The first stage of the FVV system is the capture of video streams
from the scene. Stemming from the low-cost requirement, we evaluated
several consumer depth cameras, but most of these devices project
infrared light patterns. While this is perfectly adequate for a single
camera, whenever multiple such cameras operate simultaneously, their
infrared light patterns projected at the same time on the scene interfere
with one another, which drastically decreases the quality of the depth
estimation.

Consequently, we chose the Stereolabs ZED cameras, which estimate depth
thanks to a purely passive approach, and therefore can be used together
with any number of them at the same time. Figure~\ref{fig:capture-setup}
shows our capture setup with nine ZED cameras. Professional multi-camera
setups typically use hardware dedicated clock sources to synchronize
the capture of frames among all cameras, but the ZED camera, which
is a consumer-grade device, offers no such facility. Each camera is
simply attached to the controlling computer via a USB 3.0 link, so
we have developed our own software synchronization scheme using timestamps
and a shared clock source, distributed using PTP (IEEE 1588-2002).

\begin{figure}[tbh]
\includegraphics[width=0.5\columnwidth]{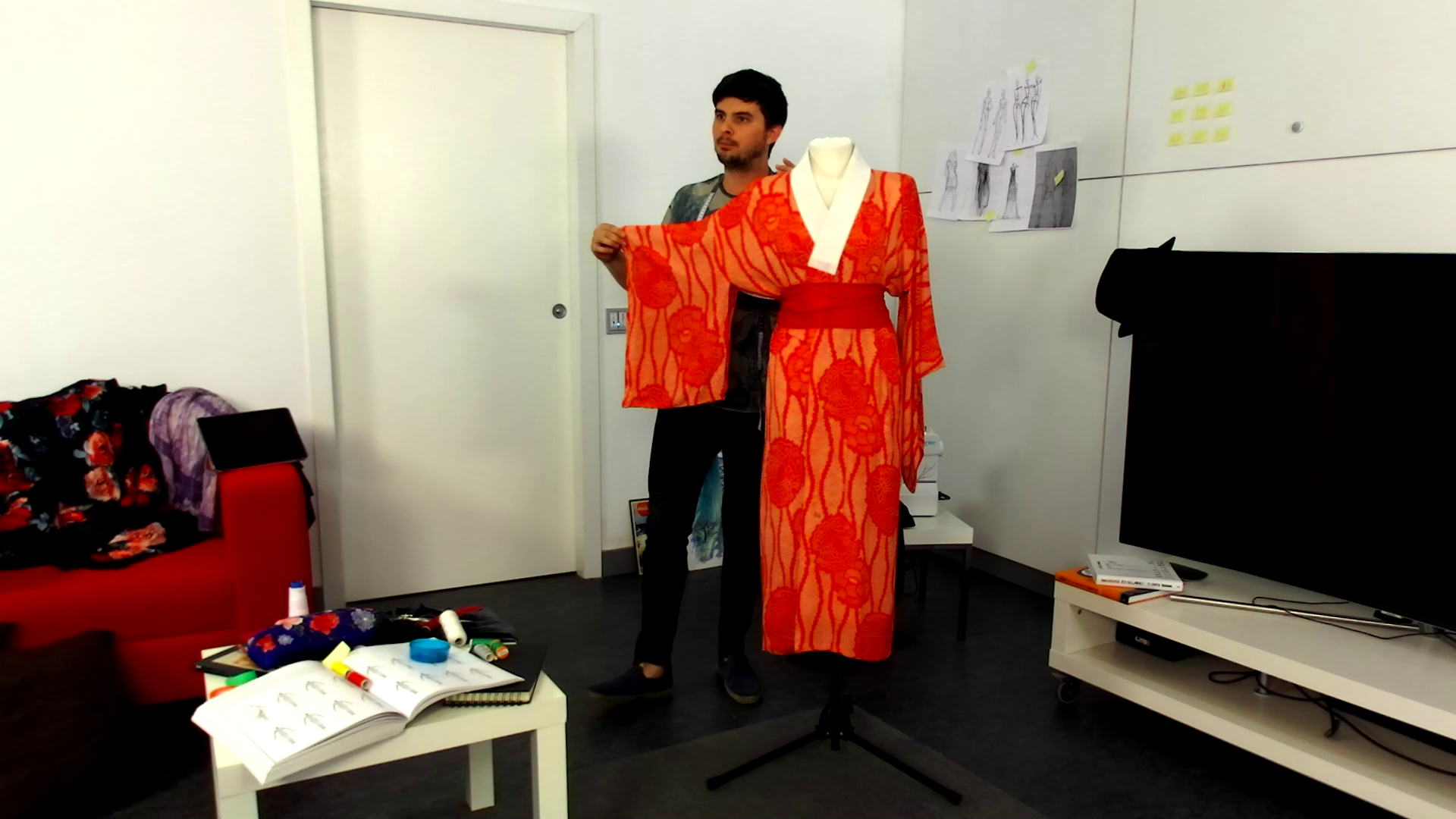}\includegraphics[width=0.5\columnwidth]{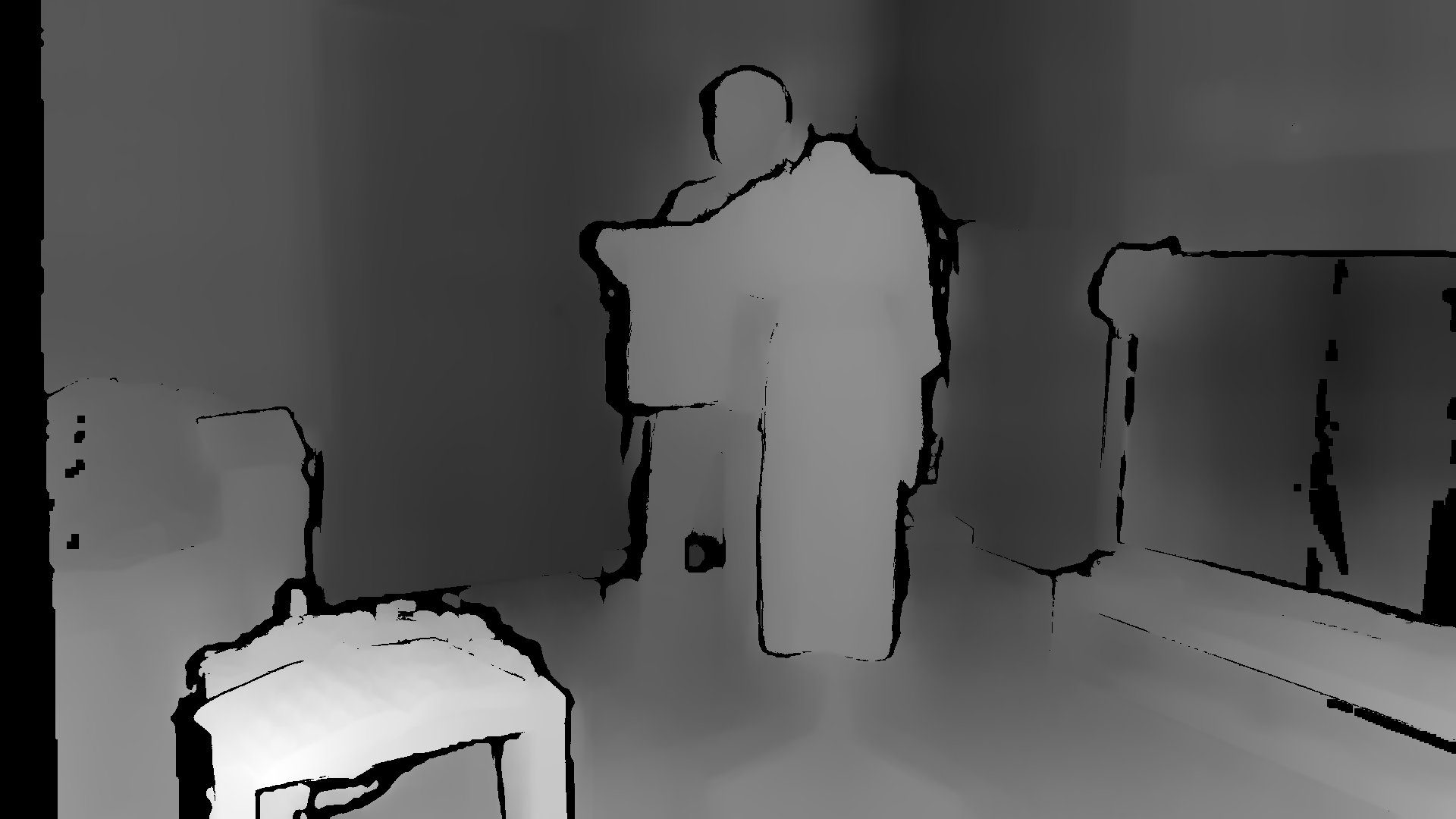}

\caption{\label{fig:sample-frame-colour-and-depth}Sample frame, colour (l)
and depth (r) information.}
\end{figure}

\begin{figure}[tbh]
\begin{centering}
\includegraphics[width=0.94\columnwidth]{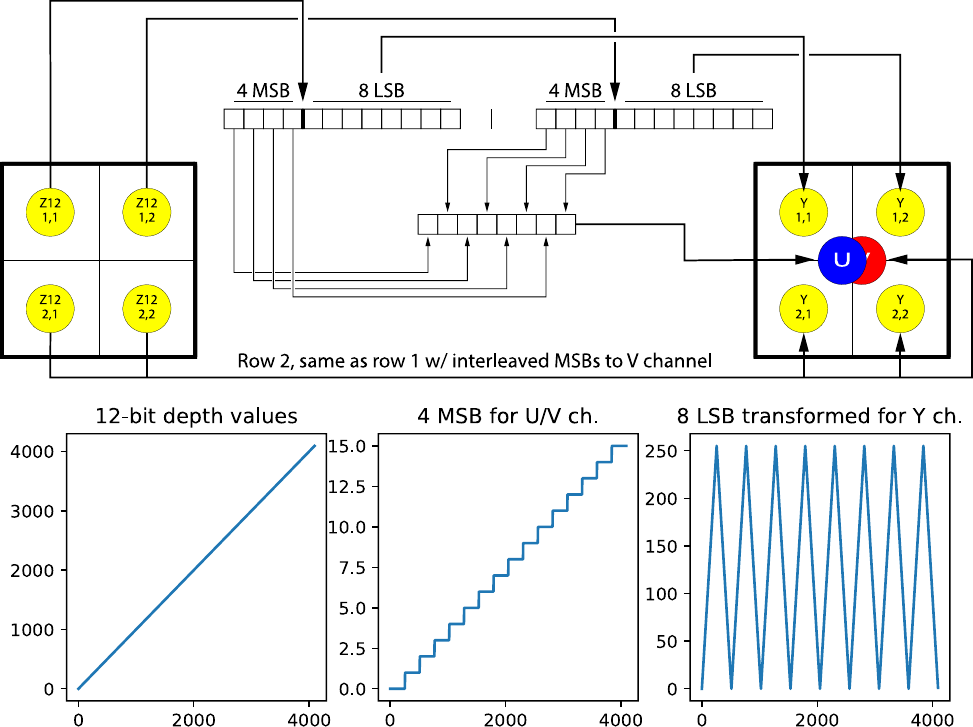}
\par\end{centering}
\caption{\label{fig:adaptation-of-420-for-depth}Adaptation of $2\times2$
pixel cells to 4:2:0 sampling structures to transport 12 bpp of depth
data.}
\end{figure}

Each camera yields regular pictures plus their corresponding depth
estimation (not actually computed by the camera, but by software on
the GPU of the controlling node), as shown in figure~\ref{fig:sample-frame-colour-and-depth}.
The sequence of regular pictures is compressed with standard (lossy)
video codecs, which cannot be blindly applied to depth data. Firstly,
more than 8 bits per pixel are required to maintain proper reconstruction
quality and, furthermore, lossy coding schemes are unacceptable because
the structure of the scene would be altered. Therefore, we have adapted
a lossless 4:2:0 video codec to transport 12 bits per pixel of depth
information by carefully arranging depth data from each $2\times2$
pixel substructure as shown in figure~\ref{fig:adaptation-of-420-for-depth}.

\pagestyle{empty}

\section{Synthesis and visualization}

\begin{figure}[tbh]
\begin{centering}
\includegraphics[width=0.9\columnwidth]{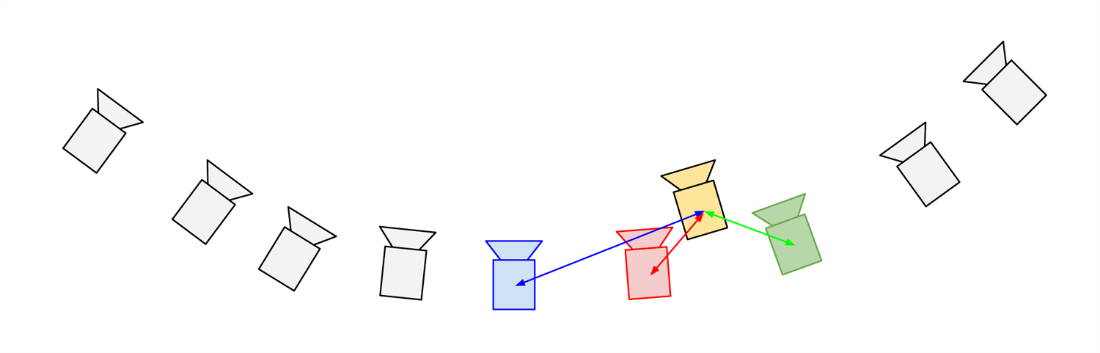}
\par\end{centering}
\caption{\label{fig:dynamic-selection-of-reference-cameras}Dynamic selection
of reference cameras for the virtual view synthesis.}
\end{figure}

A dedicated server receives video and depth streams from the capture
nodes and generates the synthetic view according to the viewpoint
and camera orientation selected by the user. In principle, if enough
computational power and network capacity were available it would be
desirable to use as much information about the scene as possible to
compute the virtual view. However, even the most powerful computers
cannot tackle problems like multi-view stereo~\cite{furukawa2010towards}
in real-time, so there is not much value in transmitting all available
data. In recognizing this fact, and considering that real-time operation
is irrenounceable for us, we dynamically select the three reference
cameras closest to the virtual viewpoint, as shown in figure~\ref{fig:dynamic-selection-of-reference-cameras},
warp them to the virtual viewpoint using DIBR techniques, and mix
their contributions to produce the synthesized view of the foreground.
Although we only use the three closest cameras to synthesize the view,
we actually transmit the five closest to ease handovers, having the
cameras that will be needed next as reference ``on call''.

To reduce the amount of data to transmit and process, we observe that
the colour of the background may change over time due to shadows or
illumination changes, but not its structure (depth). Consequently,
we can afford to generate a detailed model of the depth of the background
during system calibration (offline) using techniques that are too
costly to be used in real time (e.g., Shape from Motion, Multiview
Stereo), so that during online operation we need to send depth information
only about the foreground. Finally, we synthesize the virtual view
using a combination of layers from both background and foreground
information to produce a natural result at reduced computational cost.

\begin{figure}[tbh]
\begin{centering}
\includegraphics[width=1\columnwidth]{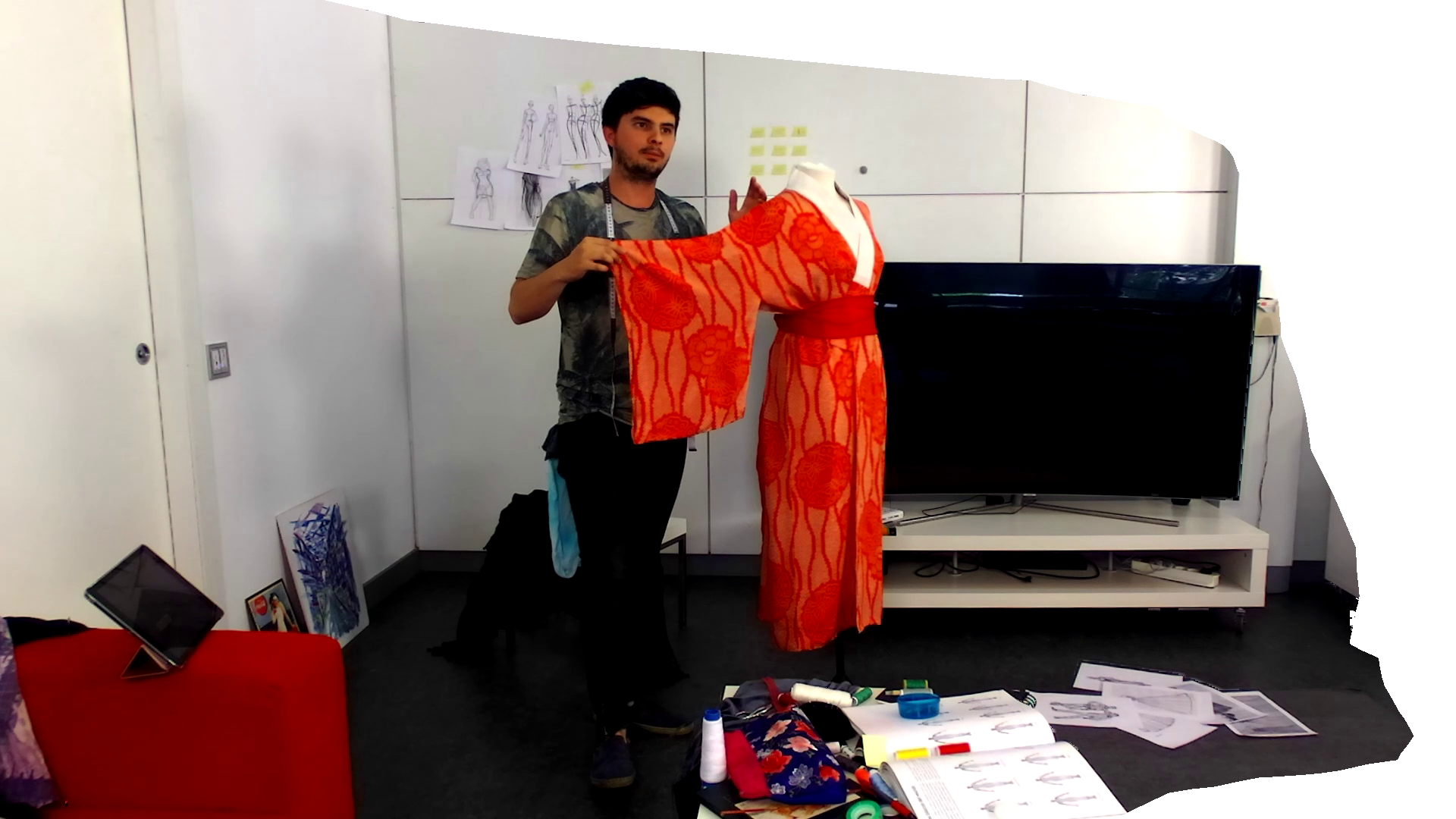}
\par\end{centering}
\caption{\label{fig:results-from-a-virtual-viewpoint}Synthetic view 
of the same instant shown in figure~\ref{fig:sample-frame-colour-and-depth}.}
\end{figure}

Thanks to its careful design and optimization, FVV Live is capable
of delivering good quality results, as shown in figure~\ref{fig:results-from-a-virtual-viewpoint}
(see \url{https://www.gti.ssr.upm.es/fvvlive} for video demos), while
maintaining real-time operation at Full HD resolution at 30~fps.

\bibliographystyle{IEEEbib}
\bibliography{fvvlive-icme2020}

\begin{thebibliography}{1}

\bibitem{trueview}
{Intel Corp.},
\newblock ``{I}ntel {T}rue {V}iew,'' [Online]:
  \url{https://web.archive.org/web/20200207150222/https://www.intel.com/content/www/us/en/sports/technology/true-view.html}.

\bibitem{4dreplay}
{4DReplay, Inc.},
\newblock ``4{D}{R}eplay,'' [Online]:
  \url{https://web.archive.org/web/20200207152030/https://www.4dreplay.com/}.

\bibitem{furukawa2010towards}
Yasutaka Furukawa, Brian Curless, Steven~M Seitz, and Richard Szeliski,
\newblock ``Towards internet-scale multi-view stereo,''
\newblock in {\em Conference on Computer Vision and Pattern Recognition}. IEEE,
  2010, pp. 1434--1441.

\end{thebibliography}

\end{document}